%% file: main.tex
\documentclass[10pt,twocolumn,letterpaper]{article}

\usepackage{iccv}              


\definecolor{iccvblue}{rgb}{0.21,0.49,0.74}
\usepackage[pagebackref,breaklinks,colorlinks,allcolors=iccvblue]{hyperref}

\usepackage{algorithm}
\usepackage{algpseudocode}
\usepackage[accsupp]{axessibility}

\def\paperID{4} 
\def\confName{ICCV}
\def\confYear{2025}

\makeatletter
\patchcmd\@maketitle{\vskip\baselineskip}{{\myfigure{}\par}}{}{}
\makeatother

\newcommand\myfigure{%
\centering
    \includegraphics[width=\linewidth]{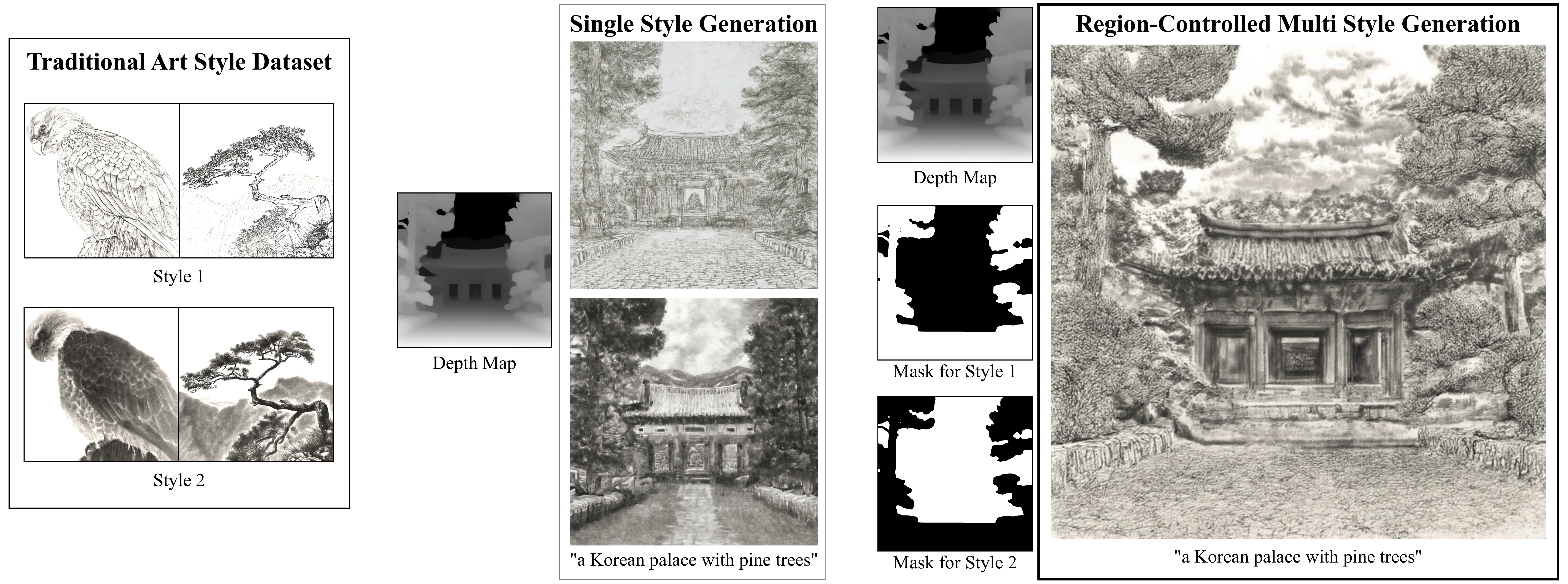}
    \captionof{figure}{Given a style dataset, we train distinct LoRA modules for each individual style. Leveraging these specialized models, our method enables regional control over multiple styles within a single image, allowing users to generate art pieces that not only reflect their intended stylistic composition but also faithfully preserve the distinctive characteristics of each style.} 
\label{fig:intro_figure}
}

\title{Style Composition within Distinct LoRA modules for Traditional Art}

\author{
  Jaehyun Lee\textsuperscript{1}\thanks{Equal contribution.} \quad
  Wonhark Park\textsuperscript{1}\footnotemark[1] \quad
  Wonsik Shin\textsuperscript{1} \quad
  Hyunho Lee\textsuperscript{1} \quad
  Hyoung Min Na\textsuperscript{2} \quad
  Nojun Kwak\textsuperscript{1} \\
  \textsuperscript{1}Seoul National University \\
  \textsuperscript{2}Kyunghee University \\
  {\footnotesize
  \textsuperscript{1}\texttt{\{jhlee795, pwh0515, wonsikshin, hhlee822, nojunk\}@snu.ac.kr}, \quad \textsuperscript{2}\texttt{leesan@khu.ac.kr}}
}

\begin{document}
\maketitle
\input{sec/0_abstract}
\input{sec/1_intro}

\input{sec/2_related_work}
\input{sec/4_method}
\input{sec/5_experiment}
\input{sec/6_conclusion_and_limitation}
{
    \small
    \bibliographystyle{ieeenat_fullname}
    \bibliography{main}
}

\end{document}

%% file: sec/0_abstract.tex
\begin{abstract}
Diffusion-based text-to-image models have achieved remarkable results in synthesizing diverse images from text prompts and can capture specific artistic styles via style personalization. However, their entangled latent space and lack of smooth interpolation make it difficult to apply distinct painting techniques in a controlled, regional manner, often causing one style to dominate. To overcome this, we propose a zero-shot diffusion pipeline that naturally blends multiple styles by performing style composition on the denoised latents predicted during the flow-matching denoising process of separately trained, style-specialized models. We leverage the fact that lower-noise latents carry stronger stylistic information and fuse them across heterogeneous diffusion pipelines using spatial masks, enabling precise, region-specific style control. This mechanism preserves the fidelity of each individual style while allowing user-guided mixing. Furthermore, to ensure structural coherence across different models, we incorporate depth-map conditioning via ControlNet into the diffusion framework. Qualitative and quantitative experiments demonstrate that our method successfully achieves region-specific style mixing according to the given masks.
\end{abstract}

%% file: sec/1_intro.tex
\section{Introduction}
\label{sec:intro}
Diffusion-based text-to-image (T2I) generative models~\cite{balaji2023ediffitexttoimagediffusionmodels,esser2024scalingrectifiedflowtransformers,nichol2022glidephotorealisticimagegeneration,podell2023sdxlimprovinglatentdiffusion, rombach2022ldm,saharia2022photorealistictexttoimagediffusionmodelsimagen,gu2022vectorquantizeddiffusionmodel} have demonstrated impressive performance, enabling the creative generation of diverse images. Notably, through T2I style transfer~\cite{ye2023ipadaptertextcompatibleimage, rout2025rbmodulation}, these models can learn image representations with specific styles and generate desired outputs based on textual prompts. In the field of art, such models allow the learning of various artistic styles, facilitating not only creative expression but also the reproduction of traditional art aesthetics.

Despite the strong generative capabilities of T2I models, applying distinct artistic techniques to image generation remains a challenge. Artistic images often rely on clearly defined painting techniques, and multiple techniques are frequently blended in complex and user-intended ways. In particular, traditional art requires adherence to specific styles while integrating multiple styles naturally. However, controlling natural blending of multiple styles is challenging due to entangled latent space in diffusion models~\cite{song2022ddim,ho2020ddpm}, unlike GANs~\cite{karras2019stylegan,kang2023scalingganstexttoimagesynthesis,li2022stylet2icompositionalhighfidelitytexttoimage,reed2016generative,xu2018attngan,li2019controllable,zhang2021cross}. As diffusion models do not support smooth interpolation in the latent space~\cite{hahm2024isometric,guo2024smooth,Peebles2022DiT,park2023understanding}, merging styles learned from separate models is difficult. A naive approach of linearly combining learned style embeddings often fails to provide controllable outputs; users cannot precisely apply specific techniques to the targeted regions, and one dominant style may overpower others. For artistic applications, it is essential to incorporate the artist's intention by allowing explicit control over where and how particular techniques are applied within the composition.

In this paper, we propose a zero-shot diffusion pipeline that enables natural style mixing, which refers applying different styles to user-specified regions (\eg applying style 1 to the certain regions and style 2 to others) in a visually coherent way without blending the styles, by leveraging the strengths of separately trained diffusion models each specialized in a specific artistic style. We observe that latents at lower noise levels in diffusion contain stronger stylistic information, and that fusion is feasible even across different diffusion pipelines. Based on this insight, we introduce a method that performs style composition on the noise-clean latents predicted during the denoising process, enabling controllable multi-style synthesis. Also, for simultaneous generation with coherent structural content in various diffusion models, we attach ControlNet~\cite{zhang2023controlnet} to the diffusion model with depth map condition.


%% file: sec/2_related_work.tex
\section{Related Work}
\label{sec:related_work}

\subsection{Flow matching}
Flow Matching~\cite{lipman2023flowmatchinggenerativemodeling} formulates generative modeling as learning a continuous-time ordinary differential equation (ODE) that transports samples from Gaussian noise to the data manifold by directly regressing a velocity field \(v_\theta(x,t)\) along a prescribed interpolation path.
Concretely, the model minimizes
\[
\mathcal{L}
= \mathbb{E}_{t\sim\mathcal{U}(0,1)}
\bigl\|v_\theta\bigl(I_t(x_0,x_1),\,t\bigr)
- \partial_t I_t(x_0,x_1)\bigr\|^2,
\]
where \(I_t(x_0,x_1)=(1-t)x_0 + t\,x_1\) linearly interpolates between noise \(x_0\) and data \(x_1\).
At inference, given the trained velocity field \(v_\theta\), samples are obtained by numerically integrating the reverse-time ODE via an explicit Euler step:
\[
z_{t_{k-1}} = z_{t_k} - \Delta t\,v_\theta(z_{t_k},t_k),
\]
where the step size \(\Delta t = t_k - t_{k-1}\) is set by partitioning \([0,1]\) into \(K\) steps, with $t_1=1$ and $t_0=0$.
More recent work integrates Flow Matching into latent diffusion pipelines~\cite{esser2024scalingrectifiedflowtransformers, flux2024} use a DiT architecture~\cite{Peebles2022DiT} to train their flow matching models, thereby enabling Flow-matching based high-quality image generation. 

\subsection{Style transfer}
Style transfer re-renders a content image in the visual style of a reference image. 
It transfers color palettes, textures, and brushwork, while preserving the original scene’s structure.
The IP-Adapter~\cite{ye2023ipadaptertextcompatibleimage} framework injects style via lightweight cross-attention adapters added in parallel to a frozen U-Net~\cite{ronneberger2015unetconvolutionalnetworksbiomedical}, capturing and transferring color, texture, and composition cues from a reference image with only ~22 M extra parameters, and remains compatible with existing control tools. 
Its plug-and-play adapters enable efficient on-the-fly style adaptation without retraining the entire diffusion network, significantly reducing computational overhead. Recently, RB-Modulation~\cite{rout2025rbmodulation} has enabled high-performance, optimization-based style transfer, dramatically reducing the required training time.

DreamBooth-LoRA~\cite{ruiz2023dreambooth, hu2022lora} specializes in style transfer by binding a dedicated style token to just a handful of reference images and using DreamBooth’s prior-preservation loss to maintain the integrity of the original content. 
Instead of fine-tuning the entire model, it updates only a compact set of low-rank adapter weights, so that each module captures the essence of a particular visual style. 
During inference, you simply swap in the appropriate LoRA~\cite{hu2022lora} module to transform any input subject into the target style, seamlessly transferring intricate artistic details while preserving the underlying scene structure. However, these methods were not inherently designed for the style-composition task of applying distinct styles to user-specified regions. Adapting them by simply masking styles onto the output image introduces unnatural artifacts and degrades overall quality. In contrast, our novel approach produces results that preserve natural coherence and visual fidelity. 

%% file: sec/4_method.tex
\begin{figure*}[htbp]
    \centering
    \includegraphics[width=\textwidth]{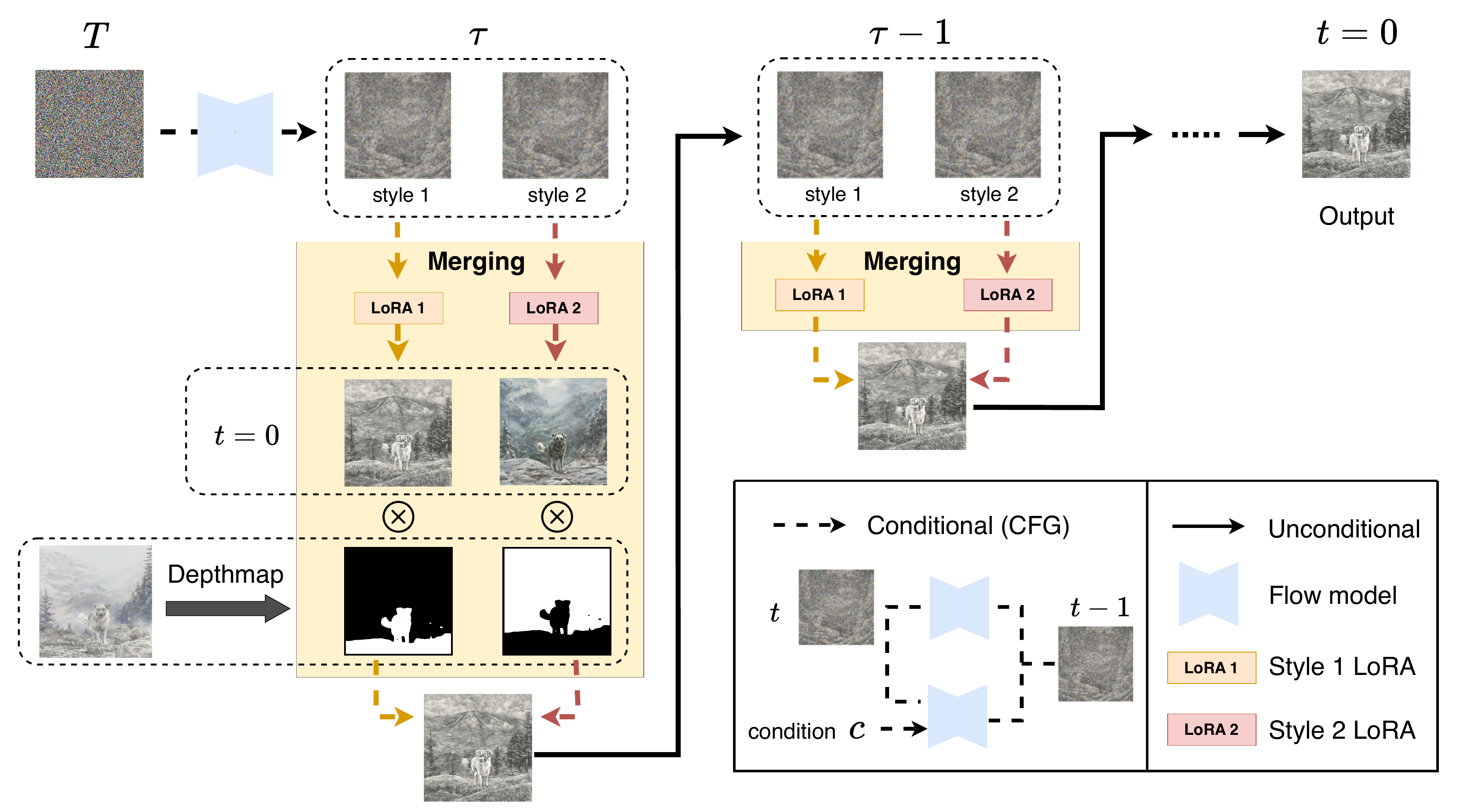}

    \caption{\textbf{Method overview.} 
    Starting from a depth map $D$, two binary masks are extracted to define the target regions. Each mask guides a separate LoRA~\cite{hu2022lora}-adapted diffusion pipeline (styles $s_1$ and $s_2$) through independent “simple updates” for coarse structure, followed by iterative “merging updates” that fuse per-style latents via Flow Matching scheduling and spatial masking. The result is a single image in which distinct styles are coherently applied to their designated regions.
    }
    \label{fig:overview}
\end{figure*}

\section{Method}
\label{sec:method}

Unlike general images, traditional artworks often exhibit highly distinctive stylistic characteristics that vary across painting styles. Some artists aim to incorporate multiple style into a single painting by regions. This objective goes beyond conventional style mixing, which blends several styles to create a novel style. Instead, our style compositive sampling targets a more controlled generation process in which distinct styles are selectively applied to the designated regions within a single image. In this paper, we address the case compositing two distinct styles based on the given depth map. The overview of the entire method is provided in Figure \ref{fig:overview}.

\subsection{Style LoRA Training}
\label{sec:training}
To enable per-style training, we employ Low-Rank Adapation (LoRA)~\cite{hu2022lora} modules, each dedicated to a single target style. Following DreamBooth~\cite{ruiz2023dreambooth}, we introduce a rare placeholder token to indicate each style, thereby guiding the LoRA training of the diffusion model. Since the joint optimization of multiple styles within a single LoRA module significantly degrades generation quality, our framework adopts a one-style-per-LoRA module strategy. In fact, since our style compositive sampling enables compositing two distinct pipelines without any optimization of fine-tuning procedure, any style personalization methods which are able to be sampled through Flow Matching~\cite{lipman2023flowmatchinggenerativemodeling} can be integrated seamlessly to our framework.

\subsection{Style Compositive Sampling}
Given depth map $D$, we aim to generate an image in which specific regions are rendered in style $s_1$, while the remaining areas are rendered in style $s_2$, thereby achieving a coherent blend of two styles. To enable this, we extract two binary masks from the depth map D by thresholding and use them to generate each style in its designated region, which are used to fuse the latents during inference. Empirically, we observe that, as timesteps progress, the attention mechanism in model induces interactions between the regions in the latent space. Consequently, employing soft masks causes style-mixing issue, whereas binary masks yield the higher-quality, more coherent outputs. We consider two diffusion models, $v_\theta^{s_1}$ and $v_\theta^{s_2}$ each equipped with a LoRA module~\cite{hu2022lora} trained to generate images in distinct styles $s_1$ and $s_2$, respectively. To incorporate structural information from depth map, we utilize ControlNet~\cite{zhang2023controlnet}, conditioned on a given depth map $D$. Specifically, we employ the DiT model~\cite{Peebles2022DiT} combined with ControlNet, and we also adopts a Flow Matching Scheduling~\cite{lipman2023flowmatchinggenerativemodeling} where a pure noise sample at time $t=1$ is denoised into an image at $t=0$ by solving ordinary differential equation (ODE). In practical, these timesteps are used discretely from $t=T\ (t=1)$ to $t=0$. For simplicity, we omit the notation for ControlNet in the followings.

The image generation process is divided into two levels, \textbf{simple updates} and \textbf{merging updates}. For the timesteps $t=T$ to $t=t'$, we apply \textbf{simple updates} that iteratively denoise each style's noise until a coarse layout becomes apparent. We delay the merge until the latent representation has acquired sufficient structural cues, thereby mitigating excessive mixing that would otherwise compromise merge fidelity. Specifically, at $t=T$, we sample pure noise and then, for each style-specific model, compute the corresponding predicted velocity to iteratively update the latent representation. This process is repeated until timestep $t'$. Note that these denoising iterations proceed completely independently for each style. 

From timestep $t'$, we perform \textbf{merging updates} by first estimating, for each style $s_i$ and current latent $z_t^{s_i}$, the conditional velocity $v^{s_i}_{t, c}=v_\theta^{s_i}(z_t^{s_i},c)$ using the style prompt $c$ and the unconditional velocity $v^{s_i}_{t, uc}=v_\theta^{s_i}(z_t^{s_i},\varnothing)$ using the null prompt $\varnothing$. We then compute the total velocity $v^{s_i}_t$ by using classifier-free guidance~\cite{ho2021classifierfree}. Leveraging the ODE formulation of Flow Matching, we then approximate the clean latent $\hat{z}^i_0$ from the current noisy latent $z^{s_i}_t$ and its velocity $v^{s_i}_t$. These per-style endpoints are fused via a spatial mask to yield the merged latent $\hat{z}_0$. To return to the timestep $t$, we propagate $\hat{z_0}$ through each unconditional velocity $v^{s_i}_{t, uc}$ for each model, which is computed ealier. Notably, since two styles co-exist in the merged latent $\hat{z_0}$, we use the unconditional velocity to avoid artifacts that would arise from conditioning on mismatched style prompts. Finally, we compute the velocity for timestep $t-1$ based on the original estimate $v^{s_i}_t$ and merge it again through mask. We iterate this process until the denoising schedule completes. 

Although latent mixing at timestep 0 was also proposed in~\cite{kwon2025tweediemix}, our problem involves blending styles rather than personalizing objects. As a result, no single prompt can reference two personalized styles simultaneously, and because the mask covers a far larger region than an object personalization, the attention mechanism exerts a far more influence on the generated images. Therefore, we had to propose an alternative novel approach as mentioned above to solve style composition task. Further details can be found in Algorithm \ref{alg:style-mixing}.

\begin{algorithm}
\caption{Style Composition via Clean Latent Fusion}
\label{alg:style-mixing}
\begin{algorithmic}[htbp]
\Statex \textbf{Input:} 
\Statex \hspace{\algorithmicindent} $i$-th Style LoRA flow-based generative models $v_\theta^{s_i}$ 
\Statex \hspace{\algorithmicindent} Style-mixing timestep $\tau$,  noise schedule $\sigma_t$
\Statex \hspace{\algorithmicindent} Source prompt embedding $c$, guidance scale $w$
\Statex \hspace{\algorithmicindent} Mask for $i$-th style $\mathcal{M}^{s_i}$
\Statex \textbf{Output:} Final style-composed image $I_{style}$

\State Sample initial noise $z_T \sim \mathcal{N}(0, I)$
\State Initialize $i$-th style latents $z_T^{s_i} = z_T$, $\forall i$
\For{$t = T$ to $\tau$}
    \State $v_t^{s_i} = v_\theta^{s_i}(z_t^{s_i},\varnothing) + w \cdot (v_\theta^{s_i}(z_t^{s_i},c) - v_\theta^{s_i}(z_t^{s_i},\varnothing))$
    \State $z_t^{s_i} = z_t^{s_i} - (\sigma_t - \sigma_{t-1}) \cdot v_t^{s_i}$
\EndFor
\For{$t = \tau - 1$ to $0$}
    \State $v_t^{s_i} = v_\theta^{s_i}(z_t^{s_i},\varnothing) + w \cdot (v_\theta^{s_i}(z_t^{s_i},c) - v_\theta^{s_i}(z_T^{s_i},\varnothing))$
    \State $\hat{z}_0^{s_i} = z_t^{s_i} - \sigma_t \cdot v_t^{s_i}$
    \State $\hat{z}_0 = \sum_i\hat{z}_0^{s_i} \cdot \mathcal{M}^{s_i}$
    \State $z_t^{s_i} = \hat{z}_0 + \sigma_t \cdot v_\theta^{s_i}(z_T^{s_i},\varnothing)$
    \State $z_t^{s_i} = z_t^{s_i} - (\sigma_t - \sigma_{t-1}) \cdot v^{s_i}$
\EndFor
\State $z_0 = \sum_i z_0^{s_i} \cdot \mathcal{M}^{s_i}$
\State $I_{style} \gets \text{Decode}(z_0)$
\end{algorithmic}
\end{algorithm}

%% file: sec/5_experiment.tex
\begin{figure*}[h]
    \centering
    \includegraphics[width=0.87\textwidth]{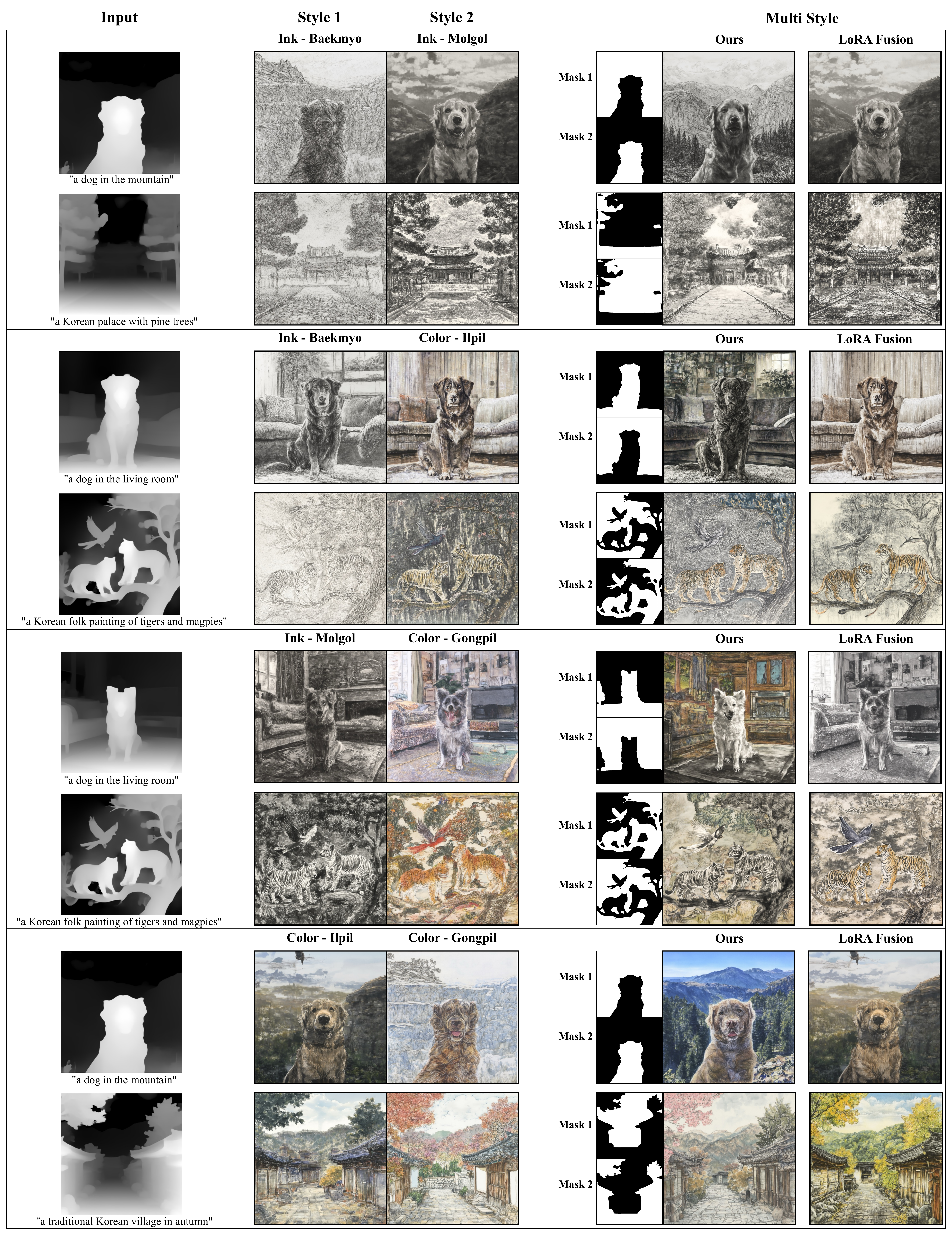}

    \caption{\textbf{Qualitative results.} 
    Each box represents a pair of styles. Given a depth map, we present the output of individual style LoRA models on the left, and the result of style mixing on the right. Our method applies each style to specific regions using separate spatial masks, enabling fine-grained control over where each style is applied. In contrast, LoRA Fusion simply averages the trained LoRA parameters, lacking spatial awareness. As a result, our approach better preserves the characteristics of each style within their designated regions while achieving more coherent and natural blending across the entire image.
    }
    \label{fig:exp_figure}
    \vspace{-0.5mm}
\end{figure*}

\section{Experiment}
\label{sec:experiment}
\subsection{Traditional Dataset}
\label{sec:dataset}
\begin{figure*}[htbp]
    \centering
    \includegraphics[width=\textwidth]{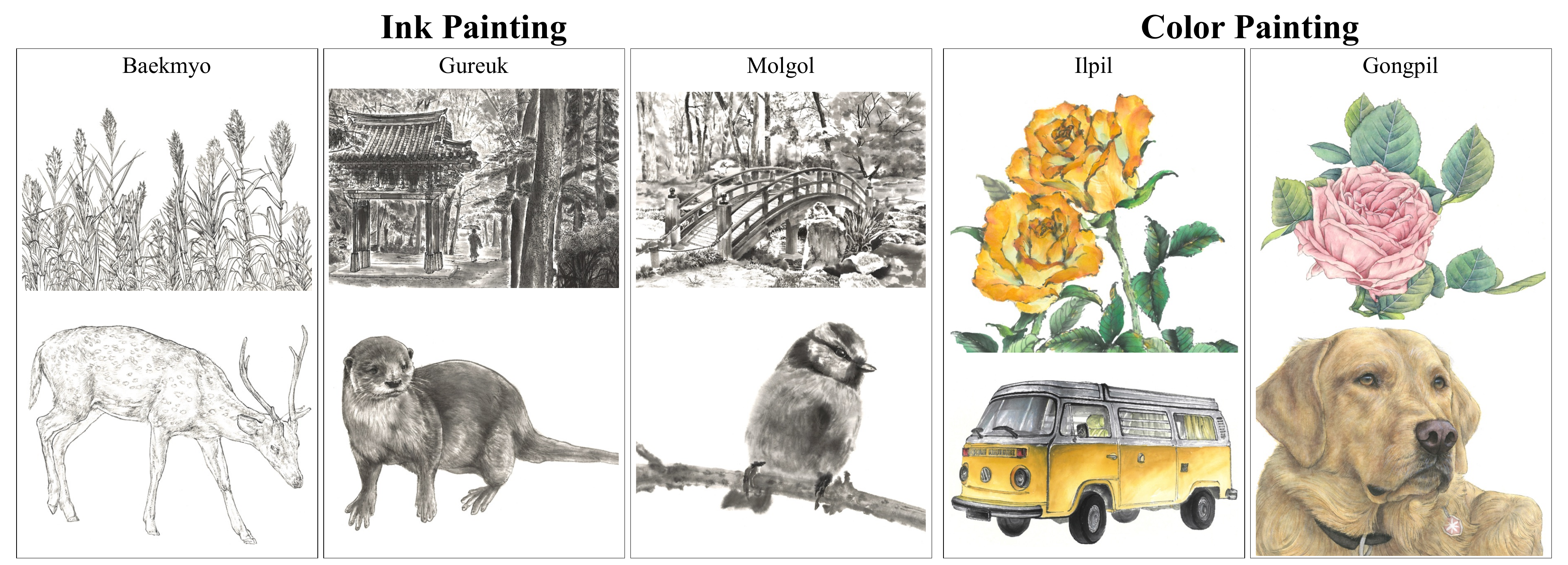}

    \caption{\textbf{Traditional Dataset.} 
    Sample images from our Korean traditional art technique dataset, consisting of five representative styles. The dataset includes three ink painting techniques (Baekmyo, Gureuk, Molgol) and two color painting techniques (Ilpil, Gongpil). Each style exhibits unique attributes in line structure, stroke pressure of the brush, and color usage. 
    }
    \label{fig:dataset_figure}
\end{figure*}

We utilize an image dataset composed of Korean traditional art drawing techniques. The dataset includes five distinct styles, each with fifty images, where three for ink painting and the other two for color painting. Examples of all five styles are presented in Figure~\ref{fig:dataset_figure}. 
 
The ink painting techniques consist of Baekmyo, Gureuk, and Molgol styles.
\begin{itemize}
    \item[$\bullet$] Baekmyo refers to technique of fine line drawing. It is a monochrome ink technique that emphasizes clean, precise outlines using fine brushwork without any coloring or shading. It captures the essence of a subject through contour lines alone, often used to depict figures or objects with clarity and restraint.
    \item[$\bullet$] Gureuk involves outlining the subject in ink and then filling the interior with color, where in ink painting, gray color. It maintains structure clarity while 
    enabling subtle expression through fillings, balancing the strength of line and the depth of hue.
    \item[$\bullet$] Molgol is a ``boneless" technique where forms are rendered directly with ink washes, without any prior outlines. It produces soft, flowing shapes that emphasize spontaneity and the natural movement of the brush.
\end{itemize}

The color painting techniques include Ilpil and Gongpil styles.  
\begin{itemize}
    \item[$\bullet$] Ilpil literally means ``one stroke," and refers to a painting style where an object or shape is expressed in a single, bold brushstroke. It emphasizes fluidity, rhythm, and the painter’s intuition, often used in expressive or abstract works.
    \item[$\bullet$] Gongpil is a meticulous coloring technique characterized by delicate lines and carefully controlled brushstrokes. It prioritizes precision, often used for highly detailed subjects such as court paintings or botanical illustrations.
\end{itemize}

\subsection{Implementation Details}
\label{implementation}
We adopted the recently proposed Flux-dev.1 model~\cite{flux2024}, which integrates DiT~\cite{Peebles2022DiT} with Flow Matching~\cite{lipman2023flowmatchinggenerativemodeling}. For LoRA~\cite{hu2022lora} adaptation, we used trained each style for 3500 steps, and set the classifier-free guidance scale to 3.5. During inference, we set the number of inference steps as 28, and initiate the \textbf{merging updates} at $t'=8$; \ie after the first 20 denoising iterations. We use threshold value as 0.5 for extracting binary masks from the given depth map in our experiments. We conduct all experiments with four style pairs, which are Baekmyo-Molgol, Baekmyo-Ilpil, Molgol-Gongpil, Ilpil-Gongpil. 

\begin{table}[h!]
\centering
\caption{CLIP Image Scores between ground truth styles and the generated images}
\begin{tabular}{cccc}
\toprule
Style1 & Style2 & LoRA Fusion & \textbf{Ours} \\
\midrule
0.673 & 0.676 & 0.692 & \textbf{0.697} \\
\bottomrule
\end{tabular}
\label{tab:quantitative}
\end{table}

\subsection{Quantitative Experiments}
\label{sec:quantitative}
We report quantitative evaluations for style 1, style 2, the naively weighted LoRA~\cite{hu2022lora} blend(LoRA Fusion), and ours for each style pairs. To measure style similarity between generated and reference images, we employ the CLIP-I scores~\cite{radford2021clip}. For both LoRA Fusion and ours, we apply spatial masks to isolate the regions for each style, compute CLIP-I scores on each masked region, and then average the two results. As reported in table~\ref{tab:quantitative}, ours show better performance than simple LoRA Fusion. We conducted experiments on the four style pairs mentioned in sec \ref{implementation}. In each pair, style1 and style 2 denote the first and the second style, respectively. The reported score is the average of the scores obtained in these four pairs.

\subsection{Qualitative Experiments}
\label{sec:qualitative}
We also conduct qualitative experiments, comparing our method with a naive weighted blending of LoRA~\cite{hu2022lora} modules (LoRA Fusion) which also mentioned in sec \ref{sec:quantitative}. Given the depth map and prompt presented in the first column, the style1 and style2 (second and third columns) illustrate the images generated by a model incorporating a single LoRA module trained on a single style for each case. Our method preserves the distinctive features of each style while allowing users to apply them to appropriate regions. To define style regions, we apply thresholding to the depth map and generate binary masks accordingly. In contrast, LoRA Fusion often suffers from style imbalance, where one style dominates the entire image or the blending process leads to the loss of distinctive characteristics associated with traditional art techniques. For instance, in 5th row (prompt "a dog in the living room" in Molgol - Gongpil case), ours constructs a coherent image in which Molgol's grayscale texture and Gongpil's color texture are distinctly presented in their respective masked regions; by contrast, in LoRA Fusion the grayscale texture dominates the entire image. Qualitative results are presented in Figure~\ref{fig:exp_figure}.

%% file: sec/6_conclusion_and_limitation.tex
\section{Conclusion and Limitation}
\label{sec:conclusion_and_limitation}

We propose a zero-shot style compositive image generation pipeline that enables regional blending of multiple artistic styles using separately trained, style-specific LoRA modules~\cite{hu2022lora}. By allowing users to explicitly assign styles to spatial regions via binary masks, our method offers fine-grained control over where and how each artistic technique is applied. Style composition is performed on the predicted noise-clean latents during each timestep of the flow matching denoising process, effectively capturing the detailed style information that each model possess. This approach not only preserves the fidelity of each individual style but also achieves smooth and coherent transitions across style boundaries. Furthermore, the modular design allows for open-ended extensibility, supporting the integration of arbitrary numbers of artistic styles without retraining a unified model.

A limitation of our current framework lies in its reliance on ControlNet~\cite{zhang2023controlnet} with a depth map condition to maintain structural consistency across style-specific diffusion models. This dependency may limit flexibility in applications where accurate depth maps are not available. \newline\newline

\par\noindent\textbf{Acknowledgements} The researchers at Seoul National University were funded by the Korean Government through the grants from NRF (2021R1A2C3006659), IITP (RS-2021-II211343) and KOCCA (RS-2024-00398320).